\pdfoutput=1
\documentclass[11pt]{article}
\usepackage{coling2020}
\usepackage{times}
\usepackage{url}
\usepackage{latexsym}
\usepackage{mathtools}
\usepackage{amssymb}
\usepackage{booktabs}
\usepackage{tabularx}
\usepackage{soul}
\usepackage{tikz}
\usepackage{pgfplots}
\usepackage{contour}
\usepackage{floatrow}
\usepackage{subcaption}

\colingfinalcopy 

\title{Automatically Identifying Words That Can Serve as
  Labels for Few-Shot Text Classification}

\author{Timo Schick \quad Helmut Schmid \quad Hinrich Sch\"utze \\[0.5em]
	Center for Information and Language Processing, LMU Munich, Germany \\[0.5em]
	{\tt schickt@cis.lmu.de}
}

\date{}

\DeclareMathOperator*{\argmax}{arg\,max}
\DeclareMathOperator*{\argmin}{arg\,min}
\DeclareMathOperator*{\avg}{avg}

\newcommand\mask{\texttt{[MASK]}}
\newcommand{\pet}{\textsc{Pet}}
\newcommand{\ours}{\textsc{Petal}}
\newcommand{\citet}{\newcite}

\setuldepth{x}

\newfloatcommand{capbtabbox}{table}[][0.96\FBwidth]

\pgfplotsset{compat=1.13}

\tikzset{
  keep name/.style={
    prefix after command={
      \pgfextra{\let\fixname\tikzlastnode}
    }
  },
  partialbox/.style={
    keep name,
    append after command={
  (\fixname.north) -- 
  (\fixname.north west) -- 
  (\fixname.south west) -- 
  ([xshift=-#1]\fixname.south)
  (\fixname.north) -- 
  (\fixname.north east) -- 
  (\fixname.south east) -- 
  ([xshift=#1]\fixname.south)
    }
  },
  partialbox/.default=5pt
}

\usetikzlibrary{calc,fit,positioning,arrows}
\pgfdeclarelayer{bg}
\pgfsetlayers{bg,main}

\definecolor{plot1}{RGB}{93,147,191}
\definecolor{plot2}{RGB}{233,72,73}
\definecolor{plot3}{RGB}{113,191,110}
\definecolor{plot4}{RGB}{163,73,151}
\definecolor{plot5}{RGB}{230,130,50}
\definecolor{decentgrey}{RGB}{180,180,180}

\newcounter{notecounter}

\newcommand{\enoteson}{\long\gdef\enote##1##2{{
\stepcounter{notecounter}
{\large\bf
\hspace{1cm}\arabic{notecounter} $<<<$ ##1: ##2
$>>>$\hspace{1cm}}}}}
\enoteson

\begin{document}
\maketitle
\begin{abstract}
A recent approach for few-shot text classification is to
convert textual inputs to cloze questions that contain some form of task description, process them with
a pretrained language model and map the predicted words to
labels. Manually defining this mapping between words and labels
requires both domain expertise and an understanding of the
language model's abilities. To mitigate this issue, we
devise an approach that automatically finds such a mapping
given small amounts of training data. For a number of tasks,
the mapping found by our approach performs almost as well as
hand-crafted label-to-word mappings.\footnote{Our implementation is publicly available at \url{https://github.com/timoschick/pet}.}
\end{abstract}

\section{Introduction}

Pretraining language models on large corpora has led to
improvements on a wide range of NLP tasks \cite[\emph{inter
    alia}]{radford2018improving,devlin2018bert,liu2019roberta},
but learning to solve tasks from only a few  examples remains a challenging problem. As small datasets are common for many real-world applications of NLP, solving this challenge is crucial to enable broad applicability. A promising direction for many tasks is to reformulate them (e.g., by appending an instruction such as ``translate into French'') so that they can directly be solved by a pretrained language model \cite{radford2018language,schick2020exploiting,brown2020language}. The key idea of \pet{} \cite{schick2020exploiting}, one such approach aimed at text classification, is to rephrase each input as a cloze question for which the language model's prediction can somehow be mapped to a label; an example is illustrated in Figure~\ref{figure:pet}.
While \pet{} achieves remarkable results with little or no labeled training data, manually defining the required mapping between a language model's predictions and labels is difficult as it requires both task-specific knowledge and an understanding of the language model's inner workings to identify words that it understands sufficiently well.

In this work, we show how this mapping can be obtained automatically, removing the need for expert knowledge: We introduce \pet{} \emph{with Automatic Labels} (\ours{}), a simple approach for identifying words that can serve as proxies for labels given small amounts of training data. At its core, our approach breaks the intractable problem of finding the mapping that maximizes the likelihood of the training data into several manageable subproblems. Integrating our approach into \pet{} significantly outperforms regular supervised training and almost matches the performance of \pet{} with a manually defined mapping.

\section{Related Work}

Reformulating problems as language modeling tasks has been explored in fully unsupervised settings \cite{radford2018language,puri2019zeroshot,davison-etal-2019-commonsense}, in few-shot scenarios with limited amounts of training data \cite{opitz2019argumentative,shwartz2020unsupervised,brown2020language}, and even in high-resource settings \cite{raffel2019exploring}. The same idea is also commonly used for probing the knowledge contained within pretrained language models \cite[\emph{inter alia}]{Petroni_2019,talmor2019olmpics,schick2019ota,ettinger2020bert}.

Our method is a direct extension of \pet{} \cite{schick2020exploiting} and is similar in spirit to \emph{automatic verbalizer search} (AVS) introduced therein. AVS is another method for automatically finding a mapping from labels to words that works as follows: First, the mapping is initialized by assigning a random word to each label and then, the mapping is improved over multiple iterations by successively replacing words with better alternatives given the current mapping in a greedy fashion. In contrast, our approach offers a closed-form solution that is conceptually simpler and faster, requires fewer hyperparameters -- which can be crucial in a data-scarce scenario -- and performs much better, especially for difficult tasks.

For \pet{}, expert knowledge is mostly encoded in the mapping from a language model's prediction to labels, which is why we focus on automating this part. The complementary problem of automatically transforming inputs \emph{before} processing them with a language model has been studied by \citet{jiang2019know}. This is also closely related to approaches for extracting patterns in relation extraction \cite{brin1999extracting,agichtein2000snowball,batista-etal-2015-semi,bouraoui2020inducing}.

\section{Pattern-Exploiting Training}

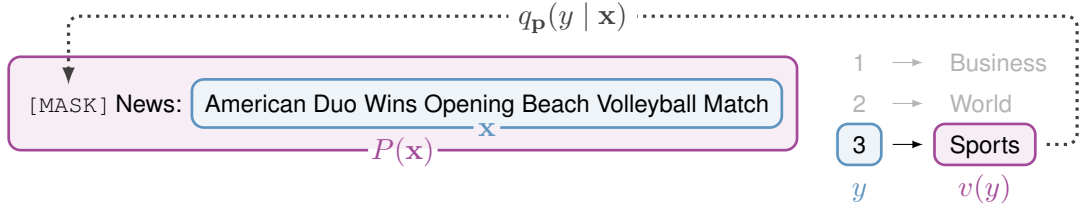
\begin{figure}
\tikzset{
  every node/.style={
  	outer sep=0, text height=1.5ex, text depth=0.25ex
  },
  input/.style={
  	draw=plot1, rounded corners, line width=2pt
  },
  pattern/.style={
  	draw=plot4, rounded corners, line width=2pt
  },
  label/.style={
  	font=\sffamily\small, rounded corners, inner ysep=0.12cm, inner xsep=0.2cm, outer xsep=0.15cm, text=decentgrey, line width=1pt
  },
  arrow/.style={
  	draw=decentgrey,->,>=latex
  },
}
\centering
\begin{tikzpicture}
\path[input] node[partialbox, font=\sffamily\small, fill=plot1!10, outer sep=0, inner sep=0.15cm, thick, align=center](input) {\textsf{American Duo Wins Opening Beach Volleyball Match}};

\node[below=0.025cm of input.south, anchor=center, outer sep=0cm, inner sep=0cm, text=plot1](input-label){ $\mathbf{x}$};

\node[font=\sffamily\small, left=0.15cm of input, inner sep=0, outer sep=0](pattern-text-1){News:};
\node[font=\sffamily\small, left=0.05cm of pattern-text-1, inner sep=0, outer ysep=0.1cm](pattern-text-2){\mask{}};

\begin{pgfonlayer}{bg}
	\path[pattern] node[partialbox=13pt, fit=(input)(pattern-text-1)(pattern-text-2), fill=plot4!10, inner ysep=0.3cm, inner xsep=0.2cm](pattern){};
	\node[below=0.025cm of pattern.south, anchor=center, outer sep=0cm, inner sep=0cm, text=plot4](pattern-label){ $P(\mathbf{x})$};
\end{pgfonlayer}

\node[label, right=0.4cm of pattern](label-2){2};
\node[label, above=0cm of label-2](label-1){1};
\node[label, below=0cm of label-2, text=black, fill=plot1!10, draw=plot1, line width=1pt](label-3){3};
\node[below=0.15cm of label-3, text=plot1, inner sep=0](y-label){$y$};

\node[label, right=0.4cm of label-2](verbalizer-2){World};
\node[label, above=0cm of verbalizer-2.north west, anchor=south west](verbalizer-1){Business};
\node[label, below=0cm of verbalizer-2.south west, anchor=north west, text=black, fill=plot4!10, draw=plot4](verbalizer-3){Sports};
\node[below=0.15cm of verbalizer-3, text=plot4, inner sep=0](y-label){$v(y)$};

\path[] (label-1) edge[arrow] (verbalizer-1);
\path[] (label-2) edge[arrow] (verbalizer-2);
\path[] (label-3) edge[arrow, draw=black] (verbalizer-3);

\draw [black!75, dotted, very thick, rounded corners, ->, >=latex] (verbalizer-3.east)--([xshift=0.4cm]verbalizer-3.east)--([xshift=0.4cm, yshift=1.7cm]verbalizer-3.east) -- ([yshift=1.7cm]verbalizer-3.east -| pattern-text-2.center) node [midway, fill=white] {$q_{\mathbf{p}}(y \mid \mathbf{x})$} -- (pattern-text-2.north);

\end{tikzpicture}
\caption{Exemplary application of a pattern-verbalizer pair $\mathbf{p} = (P,v)$: An input $\mathbf{x}$ is converted into a cloze question by applying $P$. The probability $q_\mathbf{p}(y \mid x)$ of each label $y$ is derived from the probability of its verbalization $v(y)$ being a plausible choice for the masked position.}
\label{figure:pet}
\end{figure}

We review Pattern-Exploiting Training (\pet) as proposed by
\citet{schick2020exploiting}. Let $M$ be a pretrained masked
language model (MLM),
$T$ its vocabulary and  $\mask{} \in T$ the mask token. We consider the task of mapping textual inputs $\mathbf{x} \in X$ to some label $y \in Y$ where we assume w.l.o.g. that $Y = \{1, \ldots, k \}$ for some $k \in \mathbb{N}$. 
In addition to training data $\mathcal{T} = \{ (\mathbf{x}_1, y_1), \ldots, (\mathbf{x}_n, y_n) \}$,
\pet{} requires a
set of \emph{pattern-verbalizer pairs} (PVPs). As exemplified in Figure~\ref{figure:pet}, each PVP $\mathbf{p} = (P, v)$ consists of 
\begin{itemize}
\item a \emph{pattern} $P$ that is used to convert inputs to cloze questions. Formally, $P: X \rightarrow T^*$ is defined as a function that maps each input to a sequence of tokens containing exactly one $\texttt{[MASK]}$ token;
\item a \emph{verbalizer} $v: Y \rightarrow T$ that maps each label to a single token representing its meaning. For \pet{} to work, the verbalizer must be chosen so that for each input $\mathbf{x} \in X$, $v(y)$ is a suitable replacement for the mask token in $P(\mathbf{x})$ if and only if $y$ is the correct label for $\mathbf{x}$. We call $v(y)$ the \emph{verbalization} of $y$ 
and abbreviate it as $v_y$. 
\end{itemize}
Based on this intuition, \citet{schick2020exploiting} define the conditional probability distribution $q_\mathbf{p}$ of $Y$ given $X$ as 
\begin{equation}
q_\mathbf{p}(y \mid \mathbf{x}) =  \frac{\exp M(v_y \mid P(\mathbf{x}))}{ \sum_{i=1}^k \exp M(v_i \mid P(\mathbf{x}))} \label{eq:q_p}
\end{equation}
where $M(t \mid P(\textbf{x}))$ denotes the raw score that $M$ assigns to $t$ at the masked position in $P(\mathbf{x})$; that is, the probability of $y$ being the correct label for $\mathbf{x}$ is derived from the probability of its verbalization $v_y$ being the ``correct'' token at the masked position in $P(\mathbf{x})$.

\textsc{Pet} basically works in three steps: 
\begin{enumerate}
\item For each PVP $\mathbf{p}$, a separate MLM is finetuned on $\mathcal{T}$, using the cross entropy between the true labels $y_i$ and $q_\mathbf{p}(y_i \mid \mathbf{x}_i)$ as loss function.
\item The resulting ensemble of finetuned MLMs is used to annotate a large set of unlabeled examples with soft labels.
\item Another pretrained language model with a sequence classification head is finetuned on the resulting soft-labeled dataset; this model serves as the final classifier for the task considered.
\end{enumerate}
There are several additional details to \pet{} (e.g., an additional language modeling objective to prevent catastrophic forgetting); we skip these details as they are not relevant to our approach. For a more thorough explanation, we refer to \citet{schick2020exploiting}.

\section{Likelihood Ratio Verbalizer Search}
\label{section:lrvs}

Manually defining the verbalizer $v: Y \rightarrow T$ required for \pet{} can be challenging: It requires knowledge not only of a task's labels and how they can best be expressed in natural language using a single word, but also of the used MLM's capabilities as it is crucial to choose only such words as verbalizations that are understood sufficiently well by the language model and correspond to a single token in its vocabulary.
We thus aim to automatically find a good verbalizer $v$ for some pattern $P$ without requiring task- or model-specific knowledge. 

Our method requires sets $\mathcal{V}_y \subseteq T$ of \emph{verbalization candidates} for each label $y \in Y$; for now, we simply assume $\mathcal{V}_y = T$ for all $y$. Let $\mathcal{V}$ be the set of all verbalizers consistent with these candidate sets, i.e., $v \in \mathcal{V}$ if and only if $v_y \in \mathcal{V}_y$ for all $y \in Y$. 
A natural criterion for measuring the suitability of a verbalizer $v$ is to compute the likelihood of the training data given $v$, leading to the maximum likelihood estimate
\begin{equation}
\hat{v} = \argmax_{v \in \mathcal{V} } \prod_{(\mathbf{x},y) \in \mathcal{T}} q_{(P, v)}(y \mid \mathbf{x}) \label{eq:mle}
\end{equation}
Unfortunately, iterating over $\mathcal{V}$ to find the best verbalizer is intractable: the number of possible verbalizers $|\mathcal{V}| = |T|^k$ grows exponentially in the number of labels and for a typical MLM, $T$ contains tens of thousands of tokens.

To circumvent this problem, we reframe the $k$-class
classification task as $k$ one-vs-rest classifications: For
each  $y \in Y$, we search for a verbalization $v_y$ that enables $M$ to distinguish examples with label $y$ from examples with \emph{any} other label. To this end, we introduce binarized training sets $\mathcal{T}_y = \{ (\mathbf{x}_1, \tilde{y}_1), \ldots, (\mathbf{x}_n, \tilde{y}_n) \}$ where $\tilde{y}_i = 1$ if $y_i = y$ and $0$ otherwise. For $t \in T$, we define 
\begin{equation}
q_{(P, t)}(1 \mid \mathbf{x}) =  \frac{\exp M(t \mid P(\mathbf{x}))}{\sum_{t' \in T} \exp M(t' \mid P(\mathbf{x}))}
\end{equation}
analogous  to Eq.~\ref{eq:q_p} except that we consider \emph{all} tokens $t' \in T$ for normalization,
and $q_{(P, t)}(0 \mid \mathbf{x}) = 1 - q_{(P, t)}(1 \mid \mathbf{x})$.
This enables us to formulate (and compute) the maximum likelihood estimate for each verbalization $v_y$ independently as
\begin{equation}
\hat{v}_y = \argmax_{v_y \in \mathcal{V}_y} \prod_{(\mathbf{x},\tilde{y}) \in \mathcal{T}_y} q_{(P,v_y)}(\tilde{y} \mid \mathbf{x}) \label{eq:mle-ovr}
\end{equation}
However, this reframing creates a label imbalance: If $\mathcal{T}$ is balanced, each $\mathcal{T}_y$ contains $k-1$ times as many negative examples as positive ones. To compensate for this, we raise each $q_{(P,v_y)}(\tilde{y} \mid \mathbf{x})$ to the power of 
\begin{equation}
s(\tilde{y}) = 
\begin{cases} 
1 & \text{ if } \tilde{y} = 1 \\
{n_y}/(|\mathcal{T}| - n_y) & \text{ otherwise}
\end{cases}
\end{equation}
where $n_y$ is the number of examples in $\mathcal{T}$ with label $y$. A similar fix for this imbalance problem was suggested by \citet{lee2001multicategory} for multi-class classification with support vector machines.

We next reformulate maximizing the likelihood as minimizing the cross entropy between $\tilde{y}$ and $q_{(P, v_y)}(\tilde{y}\mid \mathbf{x})$, that is, $\hat{v}_y = \argmin_{v_y \in \mathcal{V}_y} L_\text{CE}(\mathcal{T}; v_y)$ where
\begin{equation}
L_\text{CE}(\mathcal{T}; v_y) = 
-\smashoperator{\sum_{(\mathbf{x},\tilde{y}) \in \mathcal{T}_y}} s(\tilde{y}) \cdot \log q_{(P,v_y)}(\tilde{y} \mid \mathbf{x})\label{eq:celoss}
\end{equation}
This can easily be derived from Eq.~\ref{eq:mle-ovr} after compensating for the label imbalance as described above.
Unfortunately, there is the following problem with Eq.~\ref{eq:celoss}:
As the vocabulary $T$ is quite large for most pretrained MLMs, $q_{(P, v_y)}(0 \mid \mathbf{x})$ will almost always be close to $1$ and thus, $\log q_{(P, v_y)}(0 \mid \mathbf{x}) \approx \log 1 = 0$. This means that negative examples contribute almost nothing to this cross entropy loss, so optimizing for $L_\text{CE}$ results in verbalizations $\hat{v}_y$ that are \emph{overall highly likely}, but do not necessarily reflect the meaning of $y$.
We fix this problem by considering not the absolute values of $q_{(P, v_i)}(\tilde{y} \mid \mathbf{x})$, but the likelihood \emph{ratio} (LR):
\begin{equation}
L_\text{LR}(\mathcal{T}; v_y) = -\smashoperator{\sum_{(\mathbf{x},\tilde{y}) \in \mathcal{T}_y}}  s(\tilde{y}) \cdot \log
\frac{q_{(P, v_y)}(\tilde{y} \mid \mathbf{x})}{q_{(P, v_y)}(1 - \tilde{y} \mid \mathbf{x})} \label{eq:llr}
\end{equation}
Independently, this LR criterion was recently shown to compare favorably to cross entropy in gradient-based neural network training for image classification \cite{yao2020nllr}.

To arrive at $L_\text{LR}$,
we have made quite a number of modifications to our starting
point, the intractable maximum likelihood estimate.
However, the two objectives
are in fact quite similar. The key difference is that Eq.~\ref{eq:mle} enforces a large distance between $M(v_y \mid P(\mathbf{x}))$ and the \emph{maximum} score assigned to the verbalizations of other labels, whereas Eq.~\ref{eq:llr} enforces a large distance between $M(v_y \mid P(\mathbf{x}))$ and the \emph{average} score assigned to the verbalizations of other labels; this is shown in Appendix~\ref{appendix:relation-ce-lr}.

\subsection{Verbalization Candidates} 

Our above formulation requires sets of verbalization candidates $\mathcal{V}_y$ for each $y \in Y$.
These candidate sets can trivially be obtained by setting $\mathcal{V}_y = T$, but
to facilitate verbalizer search, we create candidate sets
$\mathcal{V}_y \subset T$ containing only a small subset of
the vocabulary.
First, we follow \citet{schick2020exploiting} and reduce $T$ by removing all tokens that do not correspond to real words or do not contain at least 2 alphabetic characters. From the remaining list, we collect the 10,000 tokens that occur most frequently in the task's unlabeled data and denote this filtered vocabulary by $T_f$.

As our loss formulation in Eq.~\ref{eq:llr} considers the
likelihood \emph{ratio}, it is indifferent to
the overall likelihood of a token.
To make sure that candidates
are both syntactically and semantically plausible for a
given pattern, we further restrict the set of candidates by
keeping only tokens that maximize the likelihood of all
\emph{positive examples}: For each label $y \in Y$, we
define a candidate set $T_{f,y}$ that contains the 1000
tokens $t \in T_f$ that maximize
$L_\text{CE}(\mathcal{T}_y^+; t)$ where $\mathcal{T}_y^+ =
\{ (\mathbf{x}, \tilde{y}) \in \mathcal{T}_y \mid \tilde{y}
= 1 \}$. Naturally, this induces a bias
towards frequent words. As recently shown by \citet{schick2019ota},
pretrained language models tend to understand frequent words much better than
rare words, so all other things being equal, a frequent word should be preferred over a rare word as verbalization; that is, this bias towards frequent words is indeed desirable.

\subsection{Multi-Verbalizers} 

For some tasks, it makes sense to assign multiple verbalizations to some label.\footnote{For example, one of the categories in the AG's News classification dataset \cite{zhang2015character} is ``Science/Tech'' which can best be modeled by using two verbalizations ``Science'' and ``Tech''.} This applies all the more if the verbalizations are found automatically, as it may easily occur that the most likely verbalizations for a given label cover different aspects thereof. We thus introduce the concept of \emph{multi-verbalizers}, a generalization of verbalizers to functions $v: Y \rightarrow \mathcal{P}(T)$ where $\mathcal{P}(T)$ denotes the power set of $T$. 
To integrate multi-verbalizers into \pet{}, we replace the conditional probability distribution in Eq.~\ref{eq:q_p} with
\begin{equation}
q_\mathbf{p}(y \mid \mathbf{x}) =  \frac{\exp\left(\frac{1}{|v_y|}\sum_{t \in v_y} M(t \mid P(\mathbf{x}))\right)}{ \sum_{i=1}^k \exp\left(\frac{1}{|v_i|}\sum_{t \in v_i} M(t \mid P(\mathbf{x}))\right)} \label{eq:q_p-multiverbalizer}
\end{equation}  
That is, we substitute the raw score that $M$ assigns to a label's verbalization in standard PET with the average score across all its verbalizations.

\begin{table}
\centering
\newcolumntype{Z}{>{\raggedright\arraybackslash}X}
\setlength\tabcolsep{0.2cm}
\begin{tabularx}{\linewidth}{Xlll}
\toprule
\textbf{Label} & \textbf{CE} & \textbf{LR} ($\mathcal{V}_y = T$) & \textbf{LR} ($\mathcal{V}_y = T_{f,y}$) \\
\midrule
Society
& the, The, reader
& Medieval, tradition, Biblical
& Dictionary, historical, Bible
\\
Science
& Your, the, The 
& PLoS, biomedical, phylogen 
& scientists, Physics, scientist 
\\
Health
& Your, the, reader 
& Patients, health, Health 
& health, Health, clinical 
\\
Education
& reader, Your, FAQ 
& Libraries, library, bookstore 
& library, teacher, Teachers 
\\
Computer
& reader, the, FAQ 
& toolbar, linux, gcc 
& Linux, hardware, software 
\\
Sports
& reader, Your, the 
& Racing, Motorsport, Sporting 
& sports, Sports, NASCAR 
\\
Business
& reader, Your, the
& leases, leasing, mortgages
& estate, property, finance
\\
Entertainment
& reader, Your, the
& Movie, fandom, Film
& Movie, casting, DVD
\\
Relationship
& the, reader, The
& couples, Marriage, girlfriends
& couples, Marriage, psychologist
\\
Politics
& the, The, Your
& DOJ, Constitutional, ACLU
& Constitutional, ACLU, Federal
\\
\bottomrule
\end{tabularx}
\caption{Most likely verbalizations for the Yahoo
  Questions dataset obtained using \textbf{CE} and \textbf{LR} with different candidate sets}
\label{table:verbalizers-yahoo}
\end{table}

\begin{table}
\centering
\newcolumntype{Z}{>{\raggedright\arraybackslash}X}
\setlength\tabcolsep{0.2cm}
\begin{tabularx}{\linewidth}{lZZ}
\toprule
\textbf{Label} & \textbf{AVS} ($\mathcal{V}_y = T_f$) & \textbf{LR} ($\mathcal{V}_y = T_{f,y}$) \\
\midrule
Contradiction
& insists, Kings, insist, \ul{contrary}, \ul{disagree}, Nor, Boris, maintains, Oliver, asserts
& \ul{but}, \ul{yet}, \ul{whereas}, \ul{Yet}, \ul{except}, \ul{unless}, \ul{But}, reason, unfortunately, \ul{However}
\\[0.8cm]
Neutral
& sales, Detroit, revenue, earliest, roads, artwork, designs, revenues, walls, Square
& she, he, both, god, meaning, ok, Abdul, Georgia, ad, significant
\\[0.8cm]
Entailment
& prompted, contacted, randomly, monitor, database, Register, requested, investigating, investigate, printer
& Register, Computer, \ul{Yes}, \ul{Yeah}, Alan, \ul{Sure}, \ul{Clear}, Any, Through, Howard
\\
\bottomrule
\end{tabularx}
\caption{Most likely verbalizations for the MNLI dataset obtained using \textbf{AVS} and \textbf{LR}. Suitable verbalizations are underlined.}
\label{table:verbalizers-mnli}
\end{table}

\section{Experiments}

For our experiments with \ours{}, we use the \pet{} implementation of \citet{schick2020exploiting} and follow their experimental setup. In particular, we use RoBERTa-large \cite{liu2019roberta} as underlying MLM, we use the same set of hyperparameters for \pet{}, the same evaluation tasks with the same patterns, and the same strategy for downsampling training sets. We deviate from \citet{schick2020exploiting} in that we convert all inputs to single sequences (i.e., we remove all \texttt{[SEP]} tokens) as we found this to slightly improve the verbalizers found by our approach in preliminary experiments. To ensure that our results are comparable with previous work and improvements in \pet{}'s performance are not simply due to this modification of patterns, we do so only for finding verbalizers and not for actual \pet{} training and inference.

We first analyze the verbalizers found by our method qualitatively. To this end, we consider Yahoo Questions \cite{zhang2015character}, a dataset consisting of questions and answers that have to be categorized into one of ten possible categories such as ``Health'', ``Sports'' and ``Politics''.  We use the simple pattern 
\[ P(\mathbf{x}) = \text{\mask{} Question: }\mathbf{x} \] and 50 training examples, meaning that we provide just five examples per label. Table~\ref{table:verbalizers-yahoo} shows the most likely verbalizations obtained for all labels using $L_\text{CE}$ and $L_\text{LR}$; for the latter, we consider both an unrestricted set of verbalization candidates and the candidate sets defined in Section~\ref{section:lrvs}. As can be seen, $L_\text{CE}$ does not lead to useful verbalizers for the reason outlined in Section~\ref{section:lrvs}: it only identifies words that are overall highly likely substitutes for the \mask{} in $P(\mathbf{x})$. While $L_\text{LR}$ with $\mathcal{V}_y = T$ finds reasonable verbalizers, some verbalizations are rather uncommon tokens (``PLoS'', ``phylogen'', ``gcc''); using more restrained candidate sets ($\mathcal{V}_y = T_{f,y}$) mitigates this issue and finds words that, in most instances, correspond well to the task's actual labels. The shown verbalizations also illustrate the benefit of using multi-verbalizers. For example, the verbalizations for ``Computer'' include ``hardware'' and ``software''; in isolation, none of these terms fully covers this category, but their combination does cover most of its aspects.

Next, we consider the more challenging MNLI dataset \cite{williams2018mnli}, a natural language inference dataset where given two sentences $\mathbf{x}_1$ and $\mathbf{x}_2$, the task is to decide whether both sentences contradict each other, one sentence entails the other, or neither.
On this dataset, Table~\ref{table:verbalizers-mnli} compares
\ours{} to AVS, the approach of \citet{schick2020exploiting} for automatically finding verbalizers, using the pattern \[P(\mathbf{x}_1,\mathbf{x}_2) =
\mathbf{x}_1\text{? \mask{}, 
}\mathbf{x}_2\]
and 50 labeled training examples.
While both
approaches clearly fail to find good verbalizations for the label
``Neutral'', using \ours{} results in much better
verbalizations for the other two labels, with most of the words identified by AVS being entirely unrelated to the considered labels. 

\begin{table}
\setlength\tabcolsep{0.25cm}
\begin{tabular}{lccccc}
\toprule
\multicolumn{1}{c}{\textbf{Method}}  & \multicolumn{1}{c}{\textbf{Yelp}} & \multicolumn{1}{c}{\textbf{AG's}} & \multicolumn{1}{c}{\textbf{Yahoo}} & \multicolumn{1}{c}{\textbf{MNLI}} & \multicolumn{1}{c}{\textbf{Avg.}} \\
\midrule
supervised & $44.8$ & $82.1$ & $52.5$ & $45.6$ & $56.3$ \\
\pet{} + random & $49.3$ & $83.4$ & $47.0$ & $49.2$ & $57.2$ \\
\pet{} + AVS & $55.2$ & $\mathbf{85.0}$ & $58.2$ & $52.6$ & $62.8$ \\
\ours{} (joint) & $\mathbf{56.5}$ & $84.9$ & $61.1$ & $60.9$ & $65.9$ \\
\ours{} (sep) & $55.9$ & $84.2$ & $\mathbf{62.9}$ & $\mathbf{62.4}$ & $\mathbf{66.4}$ \\
\pet{} + manual & $\underline{{60.0}}$ & $\underline{{86.3}}$ & $\underline{{66.2}}$ & $\underline{{63.9}}$ & $\underline{{69.1}}$ \\
\bottomrule
\end{tabular}
\caption{Accuracy of six methods for $|\mathcal{T}|=50$ training examples. Avg: Average across all tasks. Underlined: best overall result, bold: best result obtained without using additional task-specific knowledge}%
\label{table:overall-results}
\end{table}

To evaluate our approach quantitatively, we use the Yelp Review Full Star (Yelp) and AG's News (AG's) datasets \cite{zhang2015character} in addition to Yahoo Questions and MNLI. The task for Yelp is to guess the number of stars (ranging from 1 to 5) that a customer gave to a restaurant based on their textual review; for AG's, one of the four categories ``World'', ``Business'', ``Sports'' and ``Science/Tech'' has to be assigned to a news article.

Following \citet{schick2020exploiting}, we again consider a scenario where we have $|\mathcal{T}| = 50$ labeled training examples and a set of $10\,000 \cdot k$ unlabeled examples for each task; the unlabeled examples are only required for \pet{} and not used for finding a verbalizer. For our approach, we consider both a variant where verbalizers are computed for each pattern separately (sep), and a variant were a single verbalizer is computed for \emph{all} patterns as in AVS (joint); for the latter, the likelihood ratio losses for all patterns are simply added up and minimized jointly. We use a multi-verbalizer $\hat{v}$ where $\hat{v}(y)$ are the $n_v = 10$ most likely verbalizations per label and compare \ours{} to the following baselines:
\begin{itemize}
\item \textbf{supervised}: Regular supervised learning without \pet{}, i.e., we add a regular sequence classification head on top of the pretrained language model and perform finetuning as in \citet{devlin2018bert}.
\item \textbf{\pet{} + random}: We generate a multi-verbalizer by randomly choosing 10 words per label uniformly from $T_f$. We include this baseline to verify that any improvements over supervised learning are not simply due to \pet{} using additional unlabeled examples and auxiliary objectives, but that the actual source of improvement is the improved verbalizer.
\item \textbf{\pet{} + AVS}: We generate a multi-verbalizer with 10 labels per word using automatic verbalizer search with its default parameters.
\item \textbf{\pet{} + manual}: We consider the manually defined verbalizers of \citet{schick2020exploiting}. This serves as an upper bound of what is achievable by incorporating task- and model-specific knowledge.
\end{itemize}
Results can be seen in Table~\ref{table:overall-results}. On average, \pet{} with random verbalizers performs slightly better than regular supervised learning; we surmise that this is due to \pet{} leveraging additional unlabeled data. Random verbalizers perform much worse than AVS which, in turn, is cleary outperformed by our method for 3 out of 4 tasks, with an especially large margin on MNLI. This holds true for both the joint and sep variant of \ours{}, with the latter performing slightly better on average. Furthermore, especially for MNLI, our approach almost matches the performance of \pet{} with manually defined mappings while requiring no task-specific knowledge for finding verbalizers. The large gap between supervised learning and \ours{} is especially surprising given that the patterns -- the only other source of task-specific knowledge in \pet{} -- are very generic in nature.

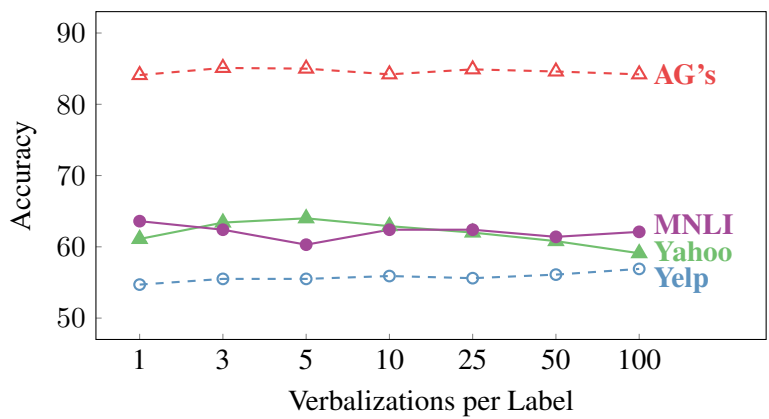
\begin{figure}
  \begin{tikzpicture}[trim axis left, trim axis right]
  \begin{axis}[
  	cycle list name=color list,
  	  xlabel={Verbalizations per Label},
  	  ylabel={Accuracy},
      ymin = 50,
      ymax = 90,
      xmin = 1,
      xmax = 8,
      enlarge x limits={0.075},
      enlarge y limits={0.075},
      xtick = {1, 2, 3, 4, 5, 6, 7},
      ytick = {50, 60, 70, 80, 90},
      xticklabels = {1, 3, 5, 10, 25, 50, 100},
      xtick pos=left,
      ytick pos=left,
      ylabel near ticks,
      xlabel near ticks,
      major tick length=0.075cm,
      width = 0.65\linewidth,
      height = 0.24\textheight,
  ]
  
   \addplot[mark=o, dashed, thick, mark options={solid}, plot1] coordinates {
   (1,54.7)
   (2,55.5)
   (3,55.5)
   (4,55.9)
   (5,55.6)
   (6,56.1)
   (7,56.9)
   } node[right,pos=1,xshift=0.05cm, yshift=-0.15cm]{{\textbf{Yelp}}};
   
   \addplot[mark=triangle, dashed, mark size=3pt, thick, mark options={solid}, plot2] coordinates {
   (1,84.1)
   (2,85.1)
   (3,85.0)
   (4,84.2)
   (5,84.9)
   (6,84.6) 
   (7,84.2)
   } node[right,pos=1,xshift=0.05cm]{{\textbf{AG's}}};

  \addplot[mark=triangle*, mark size=3pt, thick, mark options={solid}, plot3] coordinates {
  (1,61.1)
  (2,63.4)
  (3,64.0)
  (4,62.9)
  (5,62.0)
  (6,60.8)
  (7,59.1)
  } node[right,pos=1,xshift=0.05cm, yshift=0.02cm]{{\textbf{Yahoo}}};
  
  \addplot[mark=*, thick, mark options={solid}, plot4] coordinates {
  (1,63.6)
  (2,62.4)
  (3,60.3)
  (4,62.4)
  (5,62.4)
  (6,61.4)
  (7,62.1)
  } node[right,pos=1,xshift=0.05cm, yshift=0.1cm]{{\textbf{MNLI}}};
   
  \end{axis}
  \end{tikzpicture}
\caption{Performance of \ours{} (sep) on all four tasks as a function of the number of verbalizations per label ($n_v$)}%
\label{figure:num-verbalizations}
\end{figure}

We finally note that our method adds a single hyperparameter to \pet{}: the number of verbalizations per label $n_v$, which may be difficult to optimize for small training sets. However, as shown in Figure~\ref{figure:num-verbalizations}, results on all tasks are relatively stable for a wide range of values ranging from $1$ to $100$; the best result across all tasks is obtained for $n_v = 3$.
  
\section{Conclusion}

We have devised \ours{}, a simple approach that enriches \pet{} with the ability to automatically map labels to words. Qualitative and quantitative analysis shows that our approach is able to identify words that are suitable to represent labels with as little as 50 examples and almost matches the performance of hand-crafted mappings for some tasks. For future work, it would be interesting to see whether the patterns required by \pet{} can similarly be obtained in an automated fashion.

\section*{Acknowledgements}
This work was supported
by the European Research Council (grant \#740516). 

\bibliographystyle{coling}
\bibliography{literatur}

\begin{thebibliography}{}

\bibitem[\protect\citename{Agichtein and Gravano}2000]{agichtein2000snowball}
Eugene Agichtein and Luis Gravano.
\newblock 2000.
\newblock Snowball: Extracting relations from large plain-text collections.
\newblock In {\em Proceedings of the Fifth ACM Conference on Digital
  Libraries}, DL ’00, page 85–94, New York, NY, USA. Association for
  Computing Machinery.

\bibitem[\protect\citename{Batista \bgroup et al.\egroup
  }2015]{batista-etal-2015-semi}
David~S. Batista, Bruno Martins, and M{\'a}rio~J. Silva.
\newblock 2015.
\newblock Semi-supervised bootstrapping of relationship extractors with
  distributional semantics.
\newblock In {\em Proceedings of the 2015 Conference on Empirical Methods in
  Natural Language Processing}, pages 499--504, Lisbon, Portugal, September.
  Association for Computational Linguistics.

\bibitem[\protect\citename{Bouraoui \bgroup et al.\egroup
  }2020]{bouraoui2020inducing}
Zied Bouraoui, Jose Camacho-Collados, and Steven Schockaert.
\newblock 2020.
\newblock Inducing relational knowledge from {BERT}.
\newblock In {\em Proceedings of the Thirty-Fourth AAAI Conference on
  Artificial Intelligence}.

\bibitem[\protect\citename{Brin}1999]{brin1999extracting}
Sergey Brin.
\newblock 1999.
\newblock Extracting patterns and relations from the world wide web.
\newblock In Paolo Atzeni, Alberto Mendelzon, and Giansalvatore Mecca, editors,
  {\em The World Wide Web and Databases}, pages 172--183, Berlin, Heidelberg.
  Springer Berlin Heidelberg.

\bibitem[\protect\citename{Brown \bgroup et al.\egroup
  }2020]{brown2020language}
Tom~B. Brown, Benjamin Mann, Nick Ryder, Melanie Subbiah, Jared Kaplan,
  Prafulla Dhariwal, Arvind Neelakantan, Pranav Shyam, Girish Sastry, Amanda
  Askell, Sandhini Agarwal, Ariel Herbert-Voss, Gretchen Krueger, Tom Henighan,
  Rewon Child, Aditya Ramesh, Daniel~M. Ziegler, Jeffrey Wu, Clemens Winter,
  Christopher Hesse, Mark Chen, Eric Sigler, Mateusz Litwin, Scott Gray,
  Benjamin Chess, Jack Clark, Christopher Berner, Sam McCandlish, Alec Radford,
  Ilya Sutskever, and Dario Amodei.
\newblock 2020.
\newblock Language models are few-shot learners.
\newblock {\em Computing Research Repository}, arXiv:2005.14165.

\bibitem[\protect\citename{Davison \bgroup et al.\egroup
  }2019]{davison-etal-2019-commonsense}
Joe Davison, Joshua Feldman, and Alexander Rush.
\newblock 2019.
\newblock Commonsense knowledge mining from pretrained models.
\newblock In {\em Proceedings of the 2019 Conference on Empirical Methods in
  Natural Language Processing and the 9th International Joint Conference on
  Natural Language Processing (EMNLP-IJCNLP)}, pages 1173--1178, Hong Kong,
  China, November. Association for Computational Linguistics.

\bibitem[\protect\citename{Devlin \bgroup et al.\egroup }2019]{devlin2018bert}
Jacob Devlin, Ming-Wei Chang, Kenton Lee, and Kristina Toutanova.
\newblock 2019.
\newblock {BERT}: Pre-training of deep bidirectional transformers for language
  understanding.
\newblock In {\em Proceedings of the 2019 Conference of the North {A}merican
  Chapter of the Association for Computational Linguistics: Human Language
  Technologies, Volume 1 (Long and Short Papers)}, pages 4171--4186,
  Minneapolis, Minnesota, June. Association for Computational Linguistics.

\bibitem[\protect\citename{Ettinger}2020]{ettinger2020bert}
Allyson Ettinger.
\newblock 2020.
\newblock What {BERT} is not: Lessons from a new suite of psycholinguistic
  diagnostics for language models.
\newblock {\em Transactions of the Association for Computational Linguistics},
  8:34–48, Jan.

\bibitem[\protect\citename{Jiang \bgroup et al.\egroup }2019]{jiang2019know}
Zhengbao Jiang, Frank~F. Xu, Jun Araki, and Graham Neubig.
\newblock 2019.
\newblock How can we know what language models know?
\newblock {\em Computing Research Repository}, arXiv:1911.12543.

\bibitem[\protect\citename{Lee \bgroup et al.\egroup
  }2001]{lee2001multicategory}
Yoonkyung Lee, Yi~Lin, and Grace Wahba.
\newblock 2001.
\newblock Multicategory support vector machines.
\newblock Technical report, Department of Statistics, University of Madison,
  Wisconsin.

\bibitem[\protect\citename{Liu \bgroup et al.\egroup }2019]{liu2019roberta}
Yinhan Liu, Myle Ott, Naman Goyal, Jingfei Du, Mandar Joshi, Danqi Chen, Omer
  Levy, Mike Lewis, Luke Zettlemoyer, and Veselin Stoyanov.
\newblock 2019.
\newblock {RoBERTa}: {A} robustly optimized {BERT} pretraining approach.
\newblock {\em Computing Research Repository}, arXiv:1907.11692.

\bibitem[\protect\citename{Opitz}2019]{opitz2019argumentative}
Juri Opitz.
\newblock 2019.
\newblock Argumentative relation classification as plausibility ranking.
\newblock In {\em Preliminary proceedings of the 15th Conference on Natural
  Language Processing (KONVENS 2019): Long Papers}, pages 193--202, Erlangen,
  Germany. German Society for Computational Linguistics \& Language Technology.

\bibitem[\protect\citename{Petroni \bgroup et al.\egroup }2019]{Petroni_2019}
Fabio Petroni, Tim Rocktäschel, Sebastian Riedel, Patrick Lewis, Anton
  Bakhtin, Yuxiang Wu, and Alexander Miller.
\newblock 2019.
\newblock Language models as knowledge bases?
\newblock {\em Proceedings of the 2019 Conference on Empirical Methods in
  Natural Language Processing and the 9th International Joint Conference on
  Natural Language Processing (EMNLP-IJCNLP)}.

\bibitem[\protect\citename{Puri and Catanzaro}2019]{puri2019zeroshot}
Raul Puri and Bryan Catanzaro.
\newblock 2019.
\newblock Zero-shot text classification with generative language models.
\newblock {\em Computing Research Repository}, arXiv:1912.10165.

\bibitem[\protect\citename{Radford \bgroup et al.\egroup
  }2018]{radford2018improving}
Alec Radford, Karthik Narasimhan, Tim Salimans, and Ilya Sutskever.
\newblock 2018.
\newblock Improving language understanding by generative pre-training.

\bibitem[\protect\citename{Radford \bgroup et al.\egroup
  }2019]{radford2018language}
Alec Radford, Jeff Wu, Rewon Child, David Luan, Dario Amodei, and Ilya
  Sutskever.
\newblock 2019.
\newblock Language models are unsupervised multitask learners.
\newblock Technical report.

\bibitem[\protect\citename{Raffel \bgroup et al.\egroup
  }2019]{raffel2019exploring}
Colin Raffel, Noam Shazeer, Adam Roberts, Katherine Lee, Sharan Narang, Michael
  Matena, Yanqi Zhou, Wei Li, and Peter~J. Liu.
\newblock 2019.
\newblock Exploring the limits of transfer learning with a unified text-to-text
  transformer.
\newblock {\em Computing Research Repository}, arXiv:1910.10683.

\bibitem[\protect\citename{Schick and Sch{\"u}tze}2020a]{schick2020exploiting}
Timo Schick and Hinrich Sch{\"u}tze.
\newblock 2020a.
\newblock Exploiting cloze questions for few shot text classification and
  natural language inference.
\newblock {\em Computing Research Repository}, arXiv:2001.07676.

\bibitem[\protect\citename{Schick and Sch{\"u}tze}2020b]{schick2019ota}
Timo Schick and Hinrich Sch{\"u}tze.
\newblock 2020b.
\newblock Rare words: A major problem for contextualized embeddings and how to
  fix it by attentive mimicking.
\newblock In {\em Proceedings of the Thirty-Fourth AAAI Conference on
  Artificial Intelligence}.

\bibitem[\protect\citename{Shwartz \bgroup et al.\egroup
  }2020]{shwartz2020unsupervised}
Vered Shwartz, Peter West, Ronan~Le Bras, Chandra Bhagavatula, and Yejin Choi.
\newblock 2020.
\newblock Unsupervised commonsense question answering with self-talk.
\newblock {\em Computing Research Repository}, arXiv:2004.05483.

\bibitem[\protect\citename{Talmor \bgroup et al.\egroup
  }2019]{talmor2019olmpics}
Alon Talmor, Yanai Elazar, Yoav Goldberg, and Jonathan Berant.
\newblock 2019.
\newblock {oLMpics} -- on what language model pre-training captures.
\newblock {\em Computing Research Repository}, arXiv:1912.13283.

\bibitem[\protect\citename{Williams \bgroup et al.\egroup
  }2018]{williams2018mnli}
Adina Williams, Nikita Nangia, and Samuel Bowman.
\newblock 2018.
\newblock A broad-coverage challenge corpus for sentence understanding through
  inference.
\newblock In {\em Proceedings of the 2018 Conference of the North American
  Chapter of the Association for Computational Linguistics: Human Language
  Technologies, Volume 1 (Long Papers)}, pages 1112--1122. Association for
  Computational Linguistics.

\bibitem[\protect\citename{Yao \bgroup et al.\egroup }2020]{yao2020nllr}
Hengshuai Yao, Dong-lai Zhu, Bei Jiang, and Peng Yu.
\newblock 2020.
\newblock Negative log likelihood ratio loss for deep neural network
  classification.
\newblock In Kohei Arai, Rahul Bhatia, and Supriya Kapoor, editors, {\em
  Proceedings of the Future Technologies Conference (FTC) 2019}, pages
  276--282, Cham. Springer International Publishing.

\bibitem[\protect\citename{Zhang \bgroup et al.\egroup
  }2015]{zhang2015character}
Xiang Zhang, Junbo Zhao, and Yann LeCun.
\newblock 2015.
\newblock Character-level convolutional networks for text classification.
\newblock In C.~Cortes, N.~D. Lawrence, D.~D. Lee, M.~Sugiyama, and R.~Garnett,
  editors, {\em Advances in Neural Information Processing Systems 28}, pages
  649--657. Curran Associates, Inc.

\end{thebibliography}

\appendix

\section{Relation of Maximum Likelihood Estimate and One-Vs-Rest Likelihood Ratio}
\label{appendix:relation-ce-lr}

We analyze the impact of all modifications introduced in Section~\ref{section:lrvs}: reframing $k$-class classification as $k$ one-vs-rest classifications, downsampling negative examples and replacing $L_\text{CE}$ with $L_\text{LR}$. For the sake of conciseness, we drop the condition on $\mathbf{x}$ and $P(\mathbf{x})$ in $q_\mathbf{p}(y \mid \mathbf{x})$ and $M(y \mid P(\mathbf{x}))$, respectively. We start by reformulating the maximum likelihood estimate in Eq.~\ref{eq:mle} as  
\begingroup
\allowdisplaybreaks
\begin{equation}
\hat{v} = \argmin_{v \in \mathcal{V}} -\smashoperator{\sum_{(\mathbf{x}, y) \in \mathcal{T}}} \log q_{(P, v)}(y)
\end{equation}
through logarithmization and multiplication by $-1$. By applying the definition of $q_\mathbf{p}$, we obtain
\begin{align}
\hat{v} & = \argmin_{v \in \mathcal{V} } -\smashoperator{\sum_{(\mathbf{x}, y) \in \mathcal{T}}}\ \log \left( \frac{e^{M(v_y)}}{\sum_{i=1}^k e^{M(v_i)} } \right) \\
& = \argmin_{v \in \mathcal{V}} -\smashoperator{\sum_{(\mathbf{x}, y) \in \mathcal{T}}}\ \ \left( \log(e^{M(v_y)}) - \log({\sum_{y'\in Y} e^{M(v_{y'})} }) \right) \\
& = \argmin_{v \in \mathcal{V}} -\smashoperator{\sum_{(\mathbf{x}, y) \in \mathcal{T}}}\ \ \left( M(v_y) - \log({\sum_{y' \in Y} e^{M(v_{y'})} }) \right)
\end{align}
Finally, we can derive from the tangent line approximation $\log(a + b) \approx \log a + b/a$ that the left part of each addend is a soft approximation of $\max_{y' \in Y} M(v_{y'})$ (also commonly referred to as \emph{LogSumExp}), so we can approximate $\hat{v}$ as
\begin{equation}
\hat{v} \approx \argmin_{v \in \mathcal{V} } -\smashoperator{\sum_{(\mathbf{x}, y) \in \mathcal{T}}}\ \ \left( M(v_y) - \max_{y' \in Y} {M(v_{y'})} \right)
\end{equation}
\endgroup

We now consider the verbalizer obtained using $L_\text{LR}$ as in Eq.~\ref{eq:llr}, for which we assume that $\mathcal{T}$ is a balanced dataset. That is, for each label $y \in Y$, there are $|\mathcal{T}| / k$ examples with label $y$ in $\mathcal{T}$. We abbreviate the set $Y \setminus \{y\}$ of all labels except $y$ as $Y_{\setminus y}$.

As $L_\text{LR}$ for each verbalization $v_y$ is independent of all verbalizations for other labels, we can simply write the optimization criterion for $\hat{v}$ as the sum of likelihood ratio losses for all verbalizations:
\begingroup
\allowdisplaybreaks
\begin{equation}
\hat{v} = \argmin_{v \in \mathcal{V}} - \smashoperator{\sum_{y \in Y }} {\sum_{(\mathbf{x}, \tilde{y}) \in \mathcal{T}_y}} s(\tilde{y}) \cdot \log \frac{ q_{(P, v_y)}(\tilde{y}) }{ q_{(P, v_y)} (1 - \tilde{y}) } 
\end{equation}
As can be seen in the definition of $\mathcal{T}_y$, each $(\mathbf{x},y) \in \mathcal{T}$ contributes to the above sum $k$ times: $k-1$ times as negative example $(\mathbf{x}, 0) \in \mathcal{T}_{y'}$ for each $y' \neq y$, and once as a positive example $(\mathbf{x}, 1) \in \mathcal{T}_y$. We can thus rewrite the above as
\begin{align}
\hat{v} &= \argmin_{v \in \mathcal{V}} - \smashoperator{\sum_{(\mathbf{x}, y) \in \mathcal{T}}}\ \ \left( s(1) \cdot  \log \frac{ q_{(P, v_y)}(1) }{ q_{(P, v_y)} (0) } + \sum_{y' \in Y_{\setminus y}} s(0) \cdot \log \frac{ q_{(P, v_{y'})}(0) }{ q_{(P, v_{y'})} (1) }  \right)
\intertext{and again use the fact that $q_{(P, t)}(0) \approx 1$ for all $t \in T$ as well as the definition of $q_{(P,t)}$ and $s$ to obtain:}
\hat{v} &\approx \argmin_{v \in \mathcal{V}} - \smashoperator{\sum_{(\mathbf{x}, y) \in \mathcal{T}}}\ \ \left( \log  q_{(P, v_y)}(1) - \sum_{y' \in Y_{\setminus y}} s(0) \cdot \log { q_{(P, v_{y'})} (1) }  \right)
\\
&= \argmin_{v \in \mathcal{V}} - \smashoperator{\sum_{(\mathbf{x}, y) \in \mathcal{T}}}\ \ \left( 
\log \frac{e^{M(v_y)}}{\sum_{t \in T} e^{M(t)}} - \frac{1}{k-1} \sum_{y' \in Y_{\setminus y}} \log \frac{e^{M(v_{y'})}}{\sum_{t \in T} e^{M(t)}}
\right)
\end{align}
Using $\log(a / b) = \log a - \log b$ and the fact that $\sum_{t \in T} e^{M(t)}$ is independent of $v$, we can further simplify:
\begin{align}
\hat{v} &\approx \argmin_{v \in \mathcal{V}} - \smashoperator{\sum_{(\mathbf{x}, y) \in \mathcal{T}}}\ \ \left( 
\log e^{M(v_y)} - \frac{1}{k-1} \sum_{y' \in Y_{\setminus y}} \log e^{M(v_{y'})}
\right) \\
&= \argmin_{v \in \mathcal{V}} - \smashoperator{\sum_{(\mathbf{x}, y) \in \mathcal{T}}}\ \ \left( 
M(v_y) - \frac{1}{k-1} \sum_{y' \in Y_{\setminus y}} M(v_{y'}) \right) \\
&= \argmin_{v \in \mathcal{V}} - \smashoperator{\sum_{(\mathbf{x}, y) \in \mathcal{T}}}\ \ \left( 
M(v_y) - \avg_{y' \in Y_{\setminus y}} M(v_{y'}) \right)
\end{align}
\endgroup
This concludes our verification of the statement made in Section~\ref{section:lrvs}: Eq.~\ref{eq:mle} enforces a large distance between $M(v_y)$ and the \emph{maximum} score of other verbalizations, whereas Eq.~\ref{eq:llr} penalizes their \emph{average} score. 

\end{document}